\documentclass[letterpaper, 10 pt, conference]{ieeeconf}  

\IEEEoverridecommandlockouts                              

\usepackage[numbers]{natbib}
\usepackage{graphicx}
\usepackage{amsmath}
\usepackage{hyperref}
\usepackage{algpseudocode}
\usepackage{algorithm}
\usepackage[algo2e]{algorithm2e}
\usepackage{multirow}
\usepackage[normalem]{ulem}
\usepackage{booktabs}
\usepackage{gensymb}

\usepackage{amsmath,amsfonts,bm}
\usepackage{caption}
\usepackage{subcaption}

\def\eqref#1{equation~\ref{#1}}

\def\1{\bm{1}}

\DeclareMathAlphabet{\mathsfit}{\encodingdefault}{\sfdefault}{m}{sl}
\SetMathAlphabet{\mathsfit}{bold}{\encodingdefault}{\sfdefault}{bx}{n}

\newcommand{\xxnote}[3]{}
\ifx\hidenotes\undefined
  \renewcommand{\xxnote}[3]{\color{#2}{#1: #3}}
\fi

\hypersetup{
    colorlinks=true,
    linkcolor=blue,
    filecolor=magenta,      
    citecolor=blue,
    urlcolor=blue,
}

\usepackage[font={small}]{caption}
\renewcommand\thefigure{\arabic{figure}}
\title{\LARGE \bf
Dexterous Imitation Made Easy: A Learning-Based \\Framework for Efficient Dexterous Manipulation
}

\author{
Sridhar Pandian Arunachalam$^{\dagger}$, Sneha Silwal$^{\dagger}$, Ben Evans, and Lerrel Pinto\\
New York University\\
\thanks{ $\dagger$ denotes equal contribution. Correspondence to \texttt{\{sridhar, ssilwal\}@nyu.edu}.}
}

\newboolean{include-notes}
\setboolean{include-notes}{true}
\newcommand{\LP}[1]{\ifthenelse{\boolean{include-notes}}%
 {\textcolor{red}{\textbf{LP: #1}}}{}}
\newcommand{\BE}[1]{\ifthenelse{\boolean{include-notes}}%
{\textcolor{blue}{\textbf{BE: #1}}}{}}

\begin{document}

\makeatletter
\let\@oldmaketitle\@maketitle%
\renewcommand{\@maketitle}{\@oldmaketitle%
    \centering
    \includegraphics[width=\linewidth]{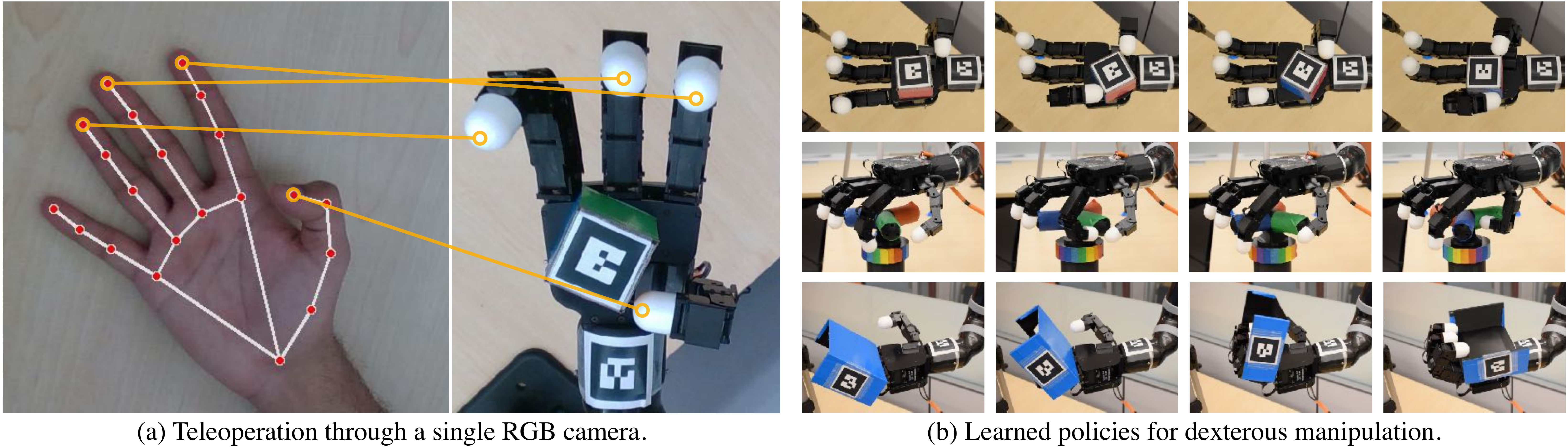}
    \captionof{figure}{The framework for dexterous manipulation consists of two phases. (a)  Demonstrations are collected using a real-time hand tracker on a single visual stream of a human operator's hand. 
    The estimated fingertip 2D pixel coordinates are retargeted to 3D coordinates in the robot frame.
    (b) Given these demonstrations,  dexterous manipulation policies are learned on both a real Allegro Hand, using nearest neighbor-based imitation, and on a simulated Allegro Hand using RL augmented with demonstrations.}
    \label{fig:intro}
}
\makeatother

\maketitle

\thispagestyle{empty}
\pagestyle{empty}

\begin{abstract}
Optimizing behaviors for dexterous manipulation has been a longstanding challenge in robotics, with a variety of methods from model-based control to model-free reinforcement learning having been previously explored in literature. Perhaps one of the most powerful techniques to learn complex manipulation strategies is imitation learning. However, collecting and learning from demonstrations in dexterous manipulation is quite challenging. The complex, high-dimensional action-space involved with multi-finger control often leads to poor sample efficiency of learning-based methods. In this work, we propose `Dexterous Imitation Made Easy' (DIME) a new imitation learning framework for dexterous manipulation. DIME only requires a single RGB camera to observe a human operator and teleoperate our robotic hand. Once demonstrations are collected, DIME employs standard imitation learning methods to train dexterous manipulation policies. On both simulation and real robot benchmarks we demonstrate that DIME can be used to solve complex, in-hand manipulation tasks such as `flipping', `spinning', and `rotating' objects with the Allegro hand. Our framework along with pre-collected demonstrations is publicly available at: \url{https://nyu-robot-learning.github.io/dime}.
\end{abstract}

\setcounter{figure}{1}

\section{Introduction}

The ability to dexterously manipulate objects with multi-fingered hands has been crucial to the development of general-purpose manipulation in humans~\cite{moravec1988mind,biagiotti2004far,dollar2007sdm}. However, multi-finger control in robots requires complex contact-rich interactions, achieved through high-dimensional actions~\cite{Kumar2016,Nagabandi2019,DBLP:journals/corr/abs-2111-03043}. Due to this, most of our robots today often only employ primitive end-effectors~\cite{tai2016state}, which significantly limits their dexterity.
To address this gap in dexterity, early works focused on developing physics-informed controllers that required precise modeling of the object-hand interaction~\cite{Dogar2010, Bai2014, Andrews2013}. Since such an ability to model the world may not be present in real-world scenarios, more recently, learning-based approaches have shown promise for general-purpose dexterity.

Over the last few years, we have witnessed several impressive results in the use of large-scale reinforcement learning (RL) for dexterous manipulation. For instance, through extensive simulator modeling, domain randomization, and millions of samples of training, behaviors such as cube rotation and Rubik's cube manipulation were demonstrated on the Shadow Hand~\cite{Openai2018,Openai2019}. However, such model-free RL techniques often require manual reward design along with several weeks of training on industry-scale compute. This begs the question -- can we learn dexterous behaviors in a sample-efficient manner?

Perhaps the most sample-efficient way to learn robotic skills is imitation learning~\cite{Bakker1996, Schaal1999}. Here, given a handful of demonstrations recorded by a human operator, the robot is tasked to imitate that behavior. So why not use imitation learning for dexterous manipulation? The key challenge is that obtaining demonstrations for high-dimensional systems is quite challenging -- kinesthetic teaching~\cite{akgun2012keyframe} requires significant conditioning on the robot; custom-built cyber gloves~\cite{glovereview} are expensive; visual piloting~\cite{handa2019dexpilot} requires registration and calibration of a multi-camera system.

In this work, we present Dexterous Imitation Made Easy (DIME), a new robotic system for both collecting and learning from visual demonstrations. Given visual inputs from a single RGB camera, DIME enables human operators to teleoperate multi-fingered robotic hands that can produce high-quality demonstrations for dexterous manipulation (See Fig.~\ref{fig:intro}(a)). Unlike prior frameworks, collecting demonstrations with DIME requires minimal calibration and human training. This is achieved by using off-the-shelf hand pose detectors~\cite{zhang2020mediapipe} to obtain fingertip positions that are then retargeted and fed to fingertip controllers on our robotic hand. Across our experiments, collecting a demonstration takes around 100 seconds on average.

Given these demonstrations, DIME is then tasked with learning policies to solve desired dexterous tasks. For this, we investigate two broad settings -- imitation in simulation and imitation on robot hardware. In simulation, we show that standard model-free RL combined with imitation learning~\cite{rajeswaran2017learning} is able to train dexterous manipulation policies in about 2 days of simulated training time. While on the real robot, we demonstrate how non-parametric, nearest-neighbor learning~\cite{rob_lwr,pari2021surprising} can achieve high performance on tasks immediately (See Fig.~\ref{fig:intro}(a)). This shows that the demonstrations obtained from DIME are versatile across imitation learning paradigms in simulation and on real robots.

In summary, this paper presents DIME, a framework that enables efficient dexterous manipulation through imitation. Concisely, our three primary contributions are: (a) We have developed an easy-to-use teleoperation framework for dexterous manipulation that can be used with untrained human operators. (b) We demonstrate the demonstrations obtained from DIME are compatible with state-of-the-art imitation learning algorithms. (c) We empirically study the interaction of imitation learning techniques with DIME and successfully solve dexterous manipulation tasks such as `flipping', `spinning', and `rotating' objects. To the best of our knowledge, DIME is the first work to successfully train dexterous manipulation policies using inexpensive demonstrations.

\begin{figure*}[t]
    \centering
    \includegraphics[width=\linewidth]{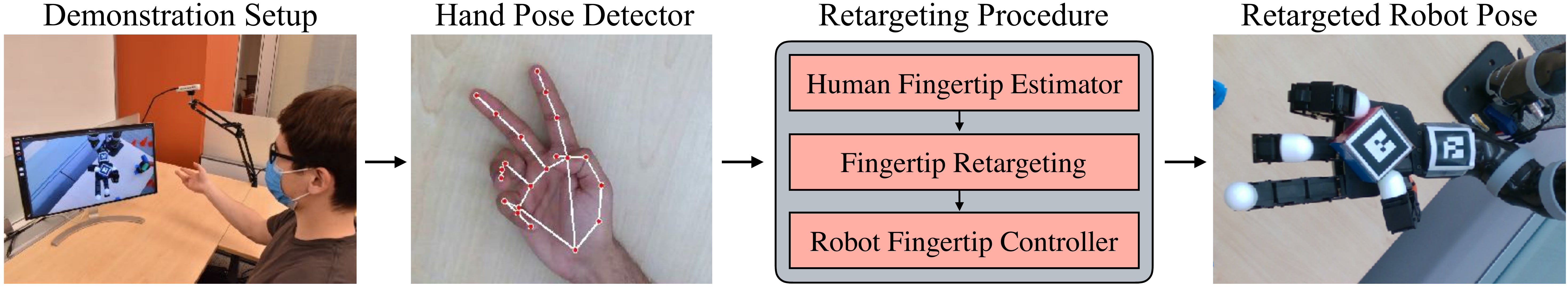}
    \caption{Overview of the teleoperation framework in DIME. Given RGB streams of a human operator's hand, a hand pose detector followed by a retargeting procedure is used to control the fingertips of the robot's hand. Visual feedback of the teleoperated actions is then provided back to the operator for real-time teleoperation.}
    \label{fig:framework}
\end{figure*}
\section{Related Work}

Our framework builds on top of several important works in collecting demonstrations, hand tracking, imitation learning, and reinforcement learning. In this section, we describe prior research that is relevant to our work.

\subsection{Obtaining Robot Demonstrations}
Many methods have been proposed to collect demonstrations, which can accelerate the learning of complex robotic tasks.
Kinesthetic training involves a human physically guiding the robot to complete a task~\cite{akgun2012keyframe,muelling2014learning,lenz2015deepmpc,sharma2018multiple}. Although this is a powerful technique for providing demonstrations to robotic arms, they are difficult to use for multi-fingered hands due to the significantly larger action space. Virtual reality-based teleoperation~\cite{zhang2018deep}, or the use of assistive tools~\cite{song2020grasping,young2020visual} have been successful for manipulation tasks. However, again they have not been shown to be useful for challenging dexterous tasks. Perhaps the most commonly used method to obtain demonstrations for dexterous hands is the use of a CyberGlove~\cite{glovereview}. Here, a custom-made glove is used to precisely measure a human operator's hand movements, which is then transferred onto a real robot. However, such a system is expensive and requires calibration before being run. More recently the DexPilot system~\cite{handa2019dexpilot} has shown how dexterous demonstrations can be produced without a Cyberglove by instead using a rig of RGBD cameras that estimate the operator's hand pose. Our framework DIME is inspired by this work and builds on top of it by alleviating the need for multiple cameras and the associated challenges of registration and calibration.

\subsection{Vision-Based Hand Tracking}
There has been a recent push in the computer vision community to detect and estimate the pose of human hands using image~\cite{handcoloredglove, zhang2020mediapipe,mobilehand} and depth-based~\cite{Schmidt2014DARTDA,v2v} observations.
Obtaining hand estimates from a single camera provides a significant advantage over expensive gloves~\cite{glovereview} or precisely-calibrated camera rigs~\cite{handa2019dexpilot}. We however note that single-camera approaches can run into unobservability due to occlusions. In our experiments, we find that our human operators avoid occlusion regions and can hence still solve dexterous tasks.
Most similar to our methodology of collecting demonstrations, recent work~\cite{sivakumar2022robotic, DBLP:journals/corr/abs-2108-05877,mandikal2021dexvip} has shown how hand-pose estimators can be used to teleoperate robotic hands. DIME builds on top of this work by collecting more dexterous demonstrations and training behavior policies using them. We use the MediaPipe hand detector~\cite{zhang2020mediapipe}, a real-time hand tracking pipeline that produces a hand skeleton from a single RGB camera. This detector is trained on a variety of hand poses across synthetic, wild, and in-house datasets to detect 21 keypoints on the hand in a 2.5D coordinate space. Coupled with GPU acceleration, this allows for more accurate and real-time pose estimates compared to other hand detectors.

\subsection{Imitation Learning}
Imitation learning is a technique to learn a policy from expert demonstrations.
Behavior Cloning (BC), encourages a model to mimic expert actions by learning over a dataset in a supervised learning fashion and has been applied to a wide range of problems \cite{ALVINN, florence2021implicit, bojarski2016end, young2020visual, young2021playful}. It is well-known that the performance of models suffers when the test data distribution shifts away from the training domain, even in the linear case~\cite{Dads}. DAgger~\cite{dagger} attempts to overcome this problem by allowing query access to an expert policy during training, something not necessarily feasible for all problems. Another approach to imitation is Inverse Reinforcement Learning (IRL), where the underlying reward function is explicitly inferred~\cite{abbeel2004apprenticeship, IRL_gt}. Following this, a policy is trained to maximize this reward function. In the context of dexterous manipulation, such IRL approaches often require extensive online training, which is not always possible for real-robotic applications.
Instead, for DIME, we use a non-parametric, nearest-neighbor approach~\cite{pari2021surprising} to map observations to actions due to its simplicity, bounded action space, and empirical success on manipulation tasks.

\subsection{Finetuning Imitation with Reinforcement Learning}
While there has been some success in using pure RL for learning policies for dexterous tasks, they are limited to simple behaviors~\cite{kumar2016optimal}, by the use of perfect models~\cite{lowrey2019plan}, or require years of simulated time for a single task~\cite{openai2019learning}. 
Incorporating external data alongside reinforcement learning algorithms has been proposed for off-task data~\cite{lynch2019learning, yarats2022dont} and with expert task-specific data by augmenting off-policy replay buffers and incorporating additional demonstration-based loss terms~\cite{vecerik2018leveraging, nair2017combining, radosavovic2020state, zhan2020framework}. Most relevant to our work is the DAPG algorithm~\cite{rajeswaran2017learning}, which has demonstrated that RL can be used to finetune behavior cloning imitation policies on a simulated Adroit hand. Similarly, we have found DAPG to be a versatile method that is able to solve imitation learning problems on a simulated Allegro hand.

\section{Imitation for Dexterity}

\subsection{Overview}
Our system, `Dexterous Imitation Made Easy' (DIME) consists of two phases -- teleoperation and imitation. In the teleoperation phase, human operators control a four-fingered Allegro hand through a visual RGB stream of their hand. This is achieved by estimating the human's hand pose, determining their fingertip locations, and retargeting them to the robot's fingertips. The robot then uses an inverse-kinematics based controller to reach desired fingertip positions. During this process, the human operator is shown real-time visual feedback of their actions. This feedback allows the operator to correct for errors and produce high-quality demonstrations for dexterous manipulation tasks.

Once a desired number of demonstrations are obtained, we begin the imitation phase. Here we investigate two settings, learning in simulation and learning in real. In the simulation setting, demonstrations are collected by teleoperating a simulated model of the Allegro hand. While in the real-robot setting, demonstrations are collected directly on the real Allegro hand. Both of these settings affords us the ability to use different learning algorithms. Simulation learning allows for sample-complex policy gradient approaches that can correct for noise in the demonstrations, while real-robot learning allows for sample-efficient non-parametric approaches. We do not run policy gradient-based learning on our real robot due to the sample complexity of such methods along with preventing accidental damage to the robot. Nevertheless, we believe that the policy gradient experiments in simulation show that DIME can be used to accelerate model-free RL as well. Details of both phases of DIME and their respective components are presented next.

\subsection{Hand Pose Mapping}
To map human hand positions to the robot, we use a single RGB camera and the MediaPipe hand detector to extract 2.5D landmarks of an operator's hand~\cite{zhang2020mediapipe}. Rather than inferring and mapping the rotational state for each joint on the human hand to the robot, we map directly from 2.5D space to 3D. Because the model does not output absolute depth estimates, we choose to ignore the depth and treat the fingertips as if they were on a plane above the palm. We then map directly from human hand fingertip locations in 2D to robot fingertip locations in 3D space, keeping the robot fingertips at a fixed height above the palm. We map the index, middle, and ring fingers along the $y$ plane.
Since the Allegro Hand only has four fingers, we use the pinky finger for finer control.
We define a quadrilateral bound with which the human thumb can be detected and re-mapped to control the $xy$ movement of the robot thumb.

Our mapping is easy to calibrate and only requires moving the hand to a sequence of reference positions at the extremes of the finger positions. We take the positions of both the human and robot hands at the extrema and linearly interpolate to produce desired target 3D positions for intermediate locations in space. The desired 3D positions are then fed to a robot controller which is described in the next section.

\subsection{Allegro Hand Controller}
Given desired fingertip locations in 3D space, we compute the required joint angles using a model of the robot and an inverse kinematics solver. For every joint, we compute the torque required to compensate for gravity and the control torque from a PD controller to reach the desired joint angle. This control loop is run at 300Hz, while desired positions are streamed at 30Hz. We use ROS~\cite{ros} to facilitate communication between the hand detector and robot. This framework is hence compatible with remote teleoperation once the remote machine is registered on our ROS network.

\begin{figure*}[t!]
    \centering
    \includegraphics[width=1\linewidth]{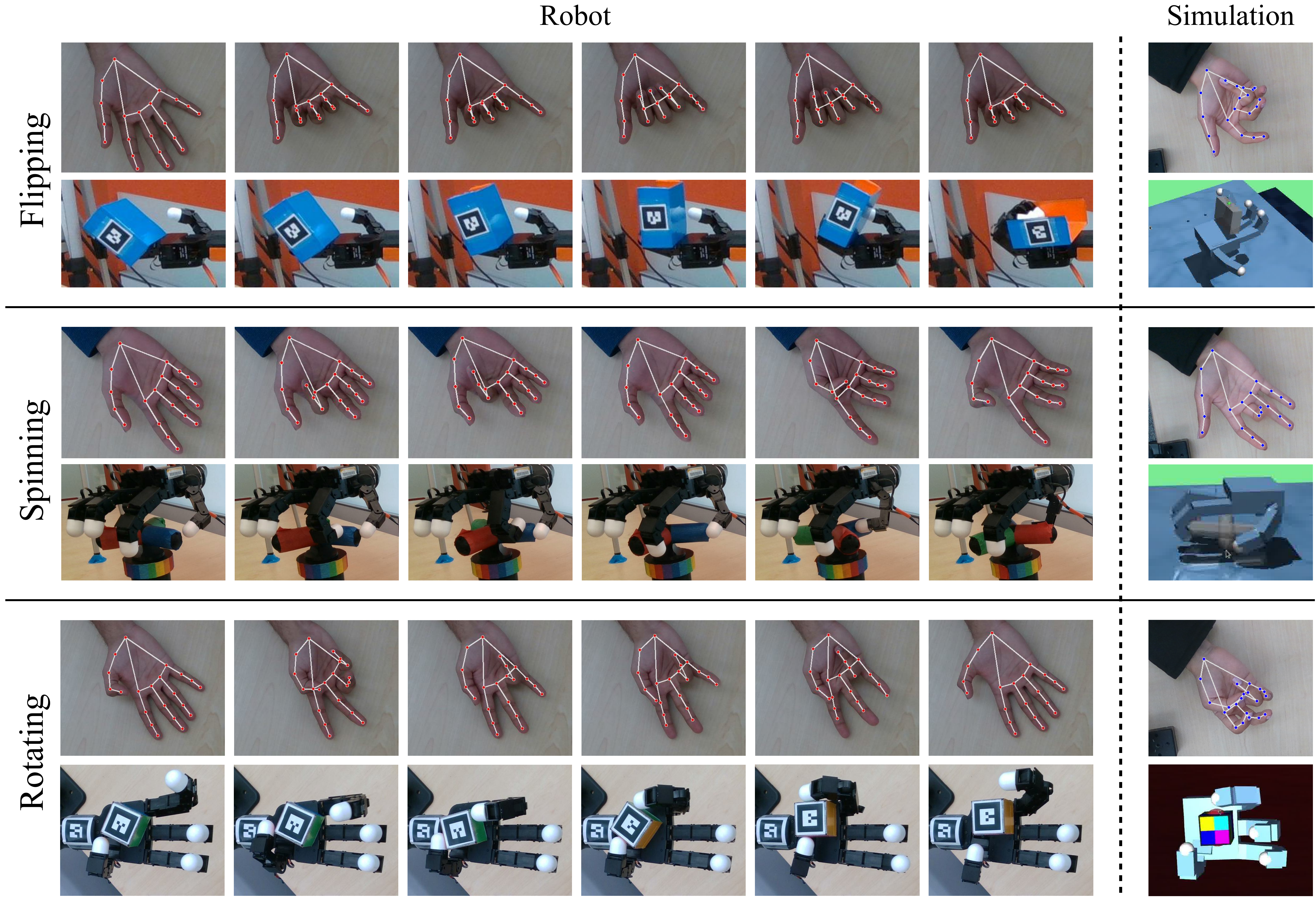}
    \caption{The demonstration collection process for the three tasks. For each task, the operator's hand in the upper row is depicted followed by the corresponding state of the robot's hand in the lower row. The rightmost column visualizes the operator's actions during demonstration collection along with the simulated MuJoCo environments for each task.}
    \label{fig:demos}
\end{figure*}

\subsection{Demonstration Collection}
To collect demonstrations, we run the hand pose detection and mapping in real-time on a single desktop computer. We record an image of the Allegro Hand, the state of the object, and the calculated target fingertip locations at 5 Hz. To estimate the state of objects we use ROS's AR tracking library. The robotic system can hence be described by the 16 joint angles of the Allegro Hand along with the 3D position and rotation of the object. For state-based imitation, we use an observation space consisting of the 3D positions of all 4 fingertips and the 3D position of the object with respect to the wrist joint. For image-based imitation, we use the RGB images of the Allegro Hand. The action space is the 3D position of the 4 desired fingertip locations, also relative to the wrist. We exclude transitions with changes less than 2 centimeters for imitation learning on the robot. This is done to account for the human operator pausing intermittently while collecting demonstrations. To study the usefulness of collected demonstrations, DIME integrates demonstration collection for both simulated hand control and real hand control. Examples of demonstrations can be seen in Fig.~\ref{fig:demos}.

Given a set of demonstrations collected through DIME, our framework for dexterous imitation studies two types of imitation learning algorithms -- RL finetuning for our simulated hand and non-parametric learning for our real hand. In the following sections we describe the algorithms.

\begin{figure*}[t!]
     \centering
     \begin{subfigure}[b]{0.34\textwidth}
         \centering
         \includegraphics[width=\textwidth]{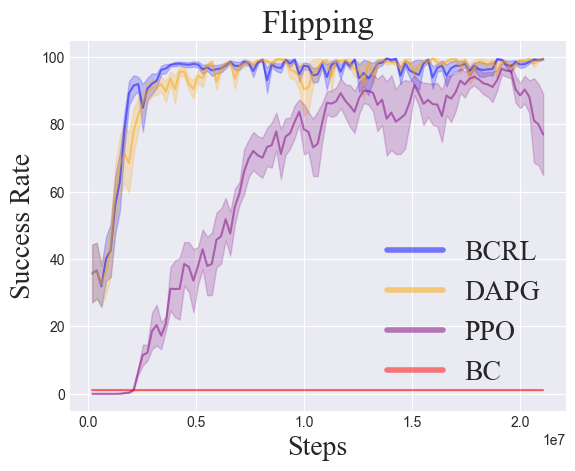}
         \label{fig:y equals x}
     \end{subfigure}
     \hfill
     \begin{subfigure}[b]{0.32\textwidth}
         \centering
         \includegraphics[width=\textwidth]{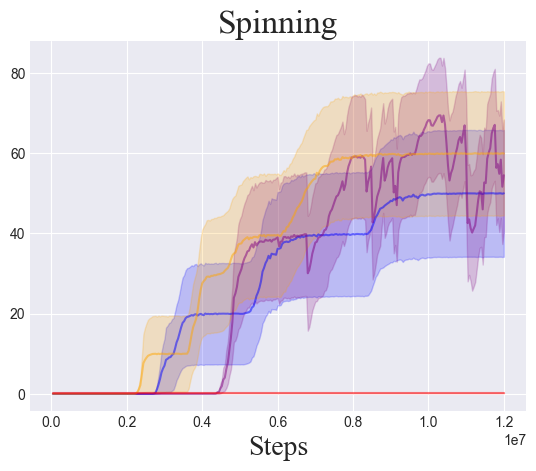}
         \label{fig:three sin x}
     \end{subfigure}
     \hfill
     \begin{subfigure}[b]{0.32\textwidth}
         \centering
         \includegraphics[width=\textwidth]{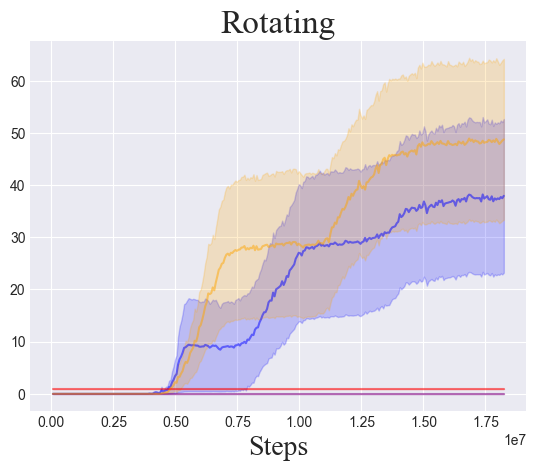}
         \label{fig:five over x}
     \end{subfigure}
    \caption{Success curves on simulated control for various RL and behavior cloning approaches with the shaded region indicating $\pm 1$ standard error measured across 10 seeds. For all tasks behavior cloning finetuned with RL (BCRL \& DAPG) achieve high success rates.}
    \label{fig:sim_rew}
\end{figure*}

\subsection{Reinforcement Learning in Simulation}
\label{sec:sim_im}
To imitate from demonstrations collected in simulation, we use Demonstration-Augmented Policy Gradients (DAPG)~\cite{rajeswaran2017learning}, a powerful imitation based algorithm for dexterous manipulation. DAPG incorporates demonstrations into the reinforcement learning procedure, which has been shown to greatly accelerate learning on a suite of difficult dexterous manipulation tasks with a 24 degree-of-freedom Adroit hand. They do so by augmenting the standard policy gradient with a weighted behavior cloning gradient term, 
\begin{equation}
\begin{split}
    g_{aug} = \sum_{(s,a) \in \rho_\pi}[\nabla_\theta \log \pi_\theta(a|s)A^\pi(s,a)] + \\ \sum_{(s,a) \in \rho_D}[\nabla_\theta \log \pi_\theta(a|s)w(s,a)]
\end{split}
\end{equation}
where $\rho_\pi, \rho_D$ are the state-action distributions generated by the policy and demonstrations, respectively, and $w(s,a)$ is an exponential weighting function that decays to zero over the course of training.
The dynamic weighting term allows the policy to quickly learn actions useful to the task and the resulting policy is able to solve tasks where standard RL fails with greater robustness to variations in mass and geometry than standard RL methods. We choose to use this method due to its simplicity and accelerated learning over prior policy-gradient approaches~\cite{vecerik2018leveraging,kakade,schulman2017proximal}.

\subsection{Imitation on the Robot}
\label{sec:rob_im}
Although DAPG can quickly learn dexterous skills in simulation, it would still require online training in the order of several days for real-robot training. Hence to study the usefulness of DIME for real robots, we turn to more sample-efficient non-parametric imitation methods.
The simplest method in this class of algorithms is $k$-nearest neighbors, which takes the current input, finds the $k$ closest inputs in the training dataset, and averages the outputs to produce a prediction~\cite{knn}. Locally weighted regression is a similar non-parametric method that weights the outputs by a similarity metric in the input space and has been used successfully for state-based~\cite{rob_lwr} (INN) and image-based~\cite{pari2021surprising} (VINN) robot learning tasks. 
For state-based nearest neighbors, we use the 15 dimensional observation space containing the 3D fingertip positions for each of the 4 fingers and the 3D position of the object.
For image-based nearest neighbors, we use 2048 dimensional feature representations of the input image obtained from an encoder trained using BYOL~\cite{grill2020bootstrap} on the images from the demonstrations~\cite{pari2021surprising}.
We find that using nearest-neighbor based imitation to select actions provides a strong baseline for dexterous manipulation tasks using a small number of demonstrations and minimal hyperparameter tuning.

\section{Experimental Evaluations}

In this section, we experimentally evaluate DIME on dexterous manipulation problems, both in simulation and on our real Allegro Hand. Our experiments seek to answer two central questions: 
\begin{itemize}
    \item Can DIME be used to collect high-quality demonstrations for dexterity?
    \item Can the produced demonstrations be used to train dexterous behaviors?
\end{itemize}

\subsection{Dexterous Manipulation Tasks}
\label{sec:tasks}
We look to tackle three dexterous manipulation tasks that reflect the challenges present in developing robot dexterity. 

\subsubsection{Flipping} Given a rectangular object placed on the fingers of the Allegro Hand, the hand is tasked to flip the object to the center of the palm. Solving this task requires precise coordination between the fingers since uneven movements result in the object falling off the hand. Performing this task is counted as a success when the object flips and lands within 2 cm of the palm's center within 1 minute.

\subsubsection{Spinning} Given a three-pronged knob attached to a table, the Allegro hand is tasked to continuously spin the knob in place. This task is based on the ROBEL benchmark~\cite{ahn2020robel}, and does not include the rotation of the knob in the observation space on hardware, but does in simulation. Performing this task is counted as a success when 120 degrees of rotation is achieved in 1 minute.

\subsubsection{Rotating} Given a cube placed in the center of the Allegro hand, the hand is tasked to continuously rotate the cube in the plane. Solving this task requires the robot to make multi-fingered contacts to both rotate and correct for deviations of the cube from the center of the hand. Performing this task is counted as a success when 90 degrees of rotation is achieved in 1 minute. 

In simulation, for each task, a MuJoCo~\cite{todorov2012mujoco} environment is created with a similar success metric as with the real robot environment. However, unlike the real environment, a dense reward function is created for each environment that is linear with the distance between the object and its target pose. This reward function is required for optimization using RL. To ensure reproducibility, our simulated environments and accompanying demonstrations will be publicly released.

\begin{figure*}[t!]
    \centering
    \includegraphics[width=\linewidth]{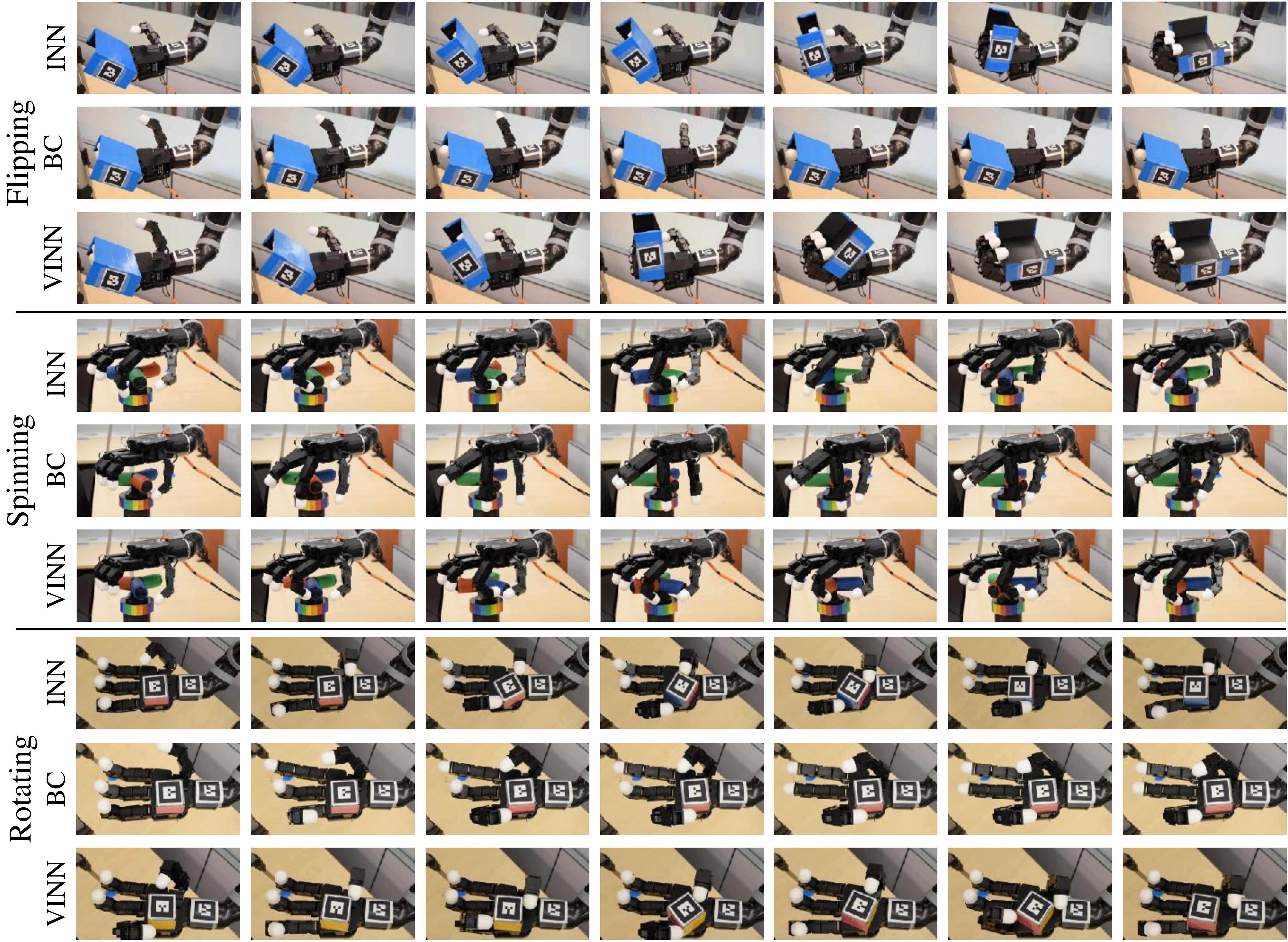}
    \caption{Robot runs for both state-based (INN) and image-based (VINN) non-parametric nearest neighbors along with state-based parametric behavior cloning (BC) are visualized across the three tasks. INN performs the best across all three tasks, while VINN is able to solve the \emph{flipping} and \emph{rotating} tasks. BC is unable to solve any task and suffers from distributional mismatch~\cite{dagger}.
    }
    \label{fig:robot_results}
\end{figure*}

\subsection{Demonstration Collection}
For each of our tasks, we first collect 30 demonstrations of each of them using DIME. Representative examples of these demonstrations can be seen in Fig.~\ref{fig:demos}. For simulation tasks, we collect 10 demonstrations for each task. On average we notice that each demonstration for \emph{flipping}, \emph{spinning}, and \emph{rotating} requires 30, 120, and 150 seconds to teleoperate respectively. This indicates that the \emph{rotating} task is the hardest among our set of tasks for human operators. We notice that since we use a single camera, there are often occlusions during teleoperation that can cause inaccuracies in the MediaPipe hand detector~\cite{zhang2020mediapipe}. To address this, human operators automatically adjust their hands through continuous visual feedback from the teleoperated robot hand.

To check if our framework for collecting demonstration is indeed easy, we asked four volunteers to collect a demonstration each for the \emph{rotating} task without any prior training on the system. We noticed that on average 210 seconds were taken to demonstrate the task on their first attempt, which is only 40\% longer than a trained expert.

\subsection{Imitation Learning Algorithms}
Given demonstrations collected in simulation and real, we study the use of the following imitation learning algorithms.

\subsubsection{Behavior Cloning (BC \& VBC)} Here, a parametric neural network model is used to predict actions given states using supervised training~\cite{Pomerleau1989,young2020visual}. We experiment with both vision-based (VBC) and state-based (BC) observations for real-robot training. 

\subsubsection{Nearest Neighbors (INN \& VINN)} Here, a non-parametric model is used to match given observations with examples in the demonstration~\cite{rob_lwr,pari2021surprising} as described in Section~\ref{sec:rob_im}. The actions corresponding to the best match are then applied to the robot. We experiment with both vision-based (VINN) and state-based (INN) observations for real-robot training.

\subsubsection{Proximal Policy Optimization (PPO)} Here, a model-free RL optimizer is used to train policies in simulation~\cite{schulman2017proximal}. Policies trained with PPO are initialized randomly.

\subsubsection{Behavior Cloning with RL finetuning (BCRL)} Here, model-free RL is used on top of a behavior-cloned policy in simulation. This can be viewed as PPO initialized from BC-trained policies.

\subsubsection{Demonstration Augmented Policy Gradient (DAPG)} Here, a policy gradient approach is used to train policies using both a cloning loss similar to BC and an RL loss similar to PPO. More details on this algorithm are presented in Section~\ref{sec:sim_im}.

The RL-based algorithms, PPO, BCRL, and DAPG are run only in simulation since running them on the real robot would require significant training time and could raise safety issues. For these experiments, we made use of the MJRL codebase~\cite{rajeswaran2017learning}. The hyperparameters for each of these algorithms are selected through a hyperparameter search. Implementations, along with tuned hyperparameters of all imitation algorithms will be publicly released to ensure reproducibility. Visualization of trained policies across all tasks can be found on our project website: \url{https://nyu-robot-learning.github.io/dime}

\subsection{Imitation Learning in Simulation}

The policies trained with BCRL and DAPG produced similar results to one another, whereas PPO methods had more erratic movements. While this erratic behavior afforded success in the environments with easier tasks (\emph{flipping} and \emph{spinning}), policies from PPO were not able to successfully rotate the cube placed on the palm. The policies that used demonstrations were also qualitatively better than the teleoperated demonstrations, which would often take longer to record and have more abrupt starts and stops. Both BCRL and DAPG policies were significantly smoother. An example of this is that the policy learned for \emph{rotating} would be able to pick up and rotate the cube between two fingers, whereas our demonstrations would keep the block on the palm and push with the thumb and last finger in opposing directions.

Surprisingly, the PPO policies were able to solve the \emph{spinning} task as well as the DAPG policies. Upon visualizing the policies, we noticed that the PPO policies were successful because extreme, random movements of the middle and last finger were enough to spin the handle.
Whereas the policies learned with demonstrations were more `human-like' and smoother. We also noted that pure behavior cloning (BC) fails on all tasks. Since the number of demonstrations used is relatively small~\cite{young2020visual}, BC policies are unable to remain in the support of demonstration data and fail~\cite{dagger}. In contrast, RL finetuning on top of the BC policies (BCRL \& PPO) allows the robot to account for states it hasn't seen in the demonstrations.

\subsection{Imitation Learning on Real Allegro Hand}

Quantitative results of our real robot experiments are presented in Table~\ref{tab:rob_results}. For each algorithm and each task, we run the robot for ten trials. Success is determined by the metrics discussed in Section~\ref{sec:tasks}. Qualitative visualization of the runs is depicted in Fig.~\ref{fig:robot_results}. We notice that non-parametric nearest neighbors (INN) outperforms parametric behavior cloning approaches across all tasks. This result is in line with recent work in non-parametric imitation~\cite{pari2021surprising} being superior to behavior cloning in domains with a limited number of demonstrations ($\sim30$). We note that on the Rotation task, which is quite difficult to solve even by human operators through teleoperation, INN achieves a perfect success rate on 90-degree rotation. Performance degrades mildly with larger rotation angles when the cube moves outside the manipulable region on the palm.

To further emphasize the usefulness of DIME, we run a visual imitation learning algorithm VINN~\cite{pari2021surprising}. The input visual information is the RGB robot images shown in Fig.~\ref{fig:robot_results}. Here, visual representations are first optimized independently for each task using the self-supervised BYOL algorithm~\cite{grill2020bootstrap}. Following this, similar to INN, nearest-neighbor matching is done to select actions. Although the performance of visual imitation is lower on average than state-based imitation, we notice strong performances on the \emph{flipping} and \emph{rotation} task. However, on the \emph{spinning} task, VINN is unable to learn good representations due to the visual complexity of the scene. We believe this is due to the inability of BYOL to learn effective representations for cluttered scenes with a limited number of demonstrations.

\begin{table}[]
\caption{Success rates on our real Allegro hand using DIME.}
\centering
\begin{tabular}{@{}ccccc@{}}
\toprule
\multirow{2}{*}{Method Used}   & \multirow{2}{*}{Flipping} & \multirow{2}{*}{Turning} & \multicolumn{2}{c}{Rotation} \\ \cmidrule(l){4-5} 
                               &                           &                          & $90^{\circ}$           & $180^{\circ}$          \\ \midrule
INN (State Based)              & 80\%                      & 60\%                     & 100\%         & 80\%         \\
Behavior Cloning (State Based) & 0\%                       & 0\%                      & 0\%           & 0\%          \\
VINN (Image Based)             & 90\%                      & 0\%                      & 70\%          & 50\%         \\
Behavior Cloning (Image Based) & 0\%                       & 0\%                      & 0\%           & 0\%          \\ \bottomrule
\end{tabular}
\label{tab:rob_results}
\end{table}


\section{Limitations and Discussion}
We have presented DIME, a framework to collect and learn from inexpensive demonstrations. Through experimental evaluations, we have shown that DIME can solve several dexterous manipulation tasks with high success rates. However, we believe that this is just the first step towards training dexterous robots from inexpensive demonstrations. There are still two key limitations of this framework. First, our demonstration collection pipeline requires the specification of a z-plane. This is because of the inherent depth ambiguity with RGB cameras. We believe this could be resolved by utilizing depth information from stereo or depth cameras. Second, for real-robot experiments, we notice that some tasks such as \emph{spinning} do not have high success rates. As evidenced by our RL experiments in simulation, we believe that developing new sample-efficient RL finetuning methods can significantly alleviate this challenge. To encourage future work on dexterous manipulation we will publicly release our custom-built hand controllers, demonstration collection pipeline, demonstration data, and learning algorithms.

\section*{Acknowledgements}
We thank David Brandfonbrener, Ilija Radosavovic, and Ankur Handa for their feedback on an early
version of the paper. We thank Balakumar Sundaralingam for help with setting up the Allegro Hand. This work was supported by a grant from Honda and ONR award number N00014-21-1-2758.


\bibliographystyle{IEEEtran}
\small
\bibliography{IEEEexample}


\end{document}